# Deep learning-based computer vision to recognize and classify suturing gestures in robot-assisted surgery

**Francisco Luongo PhD.[1], Ryan Hakim B.S. [2], Jessica H. Nguyen B.S. [2], Animashree Anandkumar PhD.[3], Andrew J. Hung M.D. [2]***

[1] Department of Biology and Biological Engineering, Caltech, Pasadena, California; [2]Center for Robotic Simulation & Education, Catherine & Joseph Aresty Department of Urology, USC Institute of Urology, University of Southern California, Los Angeles, California; [3]Department of Computing & Mathematical Sciences, Caltech, Pasadena, California

**\*Corresponding Author:**
Andrew J. Hung, MD
University of Southern California Institute of Urology
1441 Eastlake Avenue Suite 7416
Los Angeles, California 90089
Phone number:        (+1) 323-865-3700
Fax number:        (+1) 323-865-0120

**Email addresses:**        fluongo@gmail.com
        Ryanhakim3@gmail.com
        Jessica.Nguyen@med.usc.edu
        anima@caltech.edu
        Andrew.Hung@med.usc.edu

**Article Summary:** This study utilizes deep learning-based computer vision to identify specific moments of surgical suturing activity and to classify each specific suturing gesture applied during robot-assisted surgery. The importance of this work is the foundation it provides for future automation of surgical skill assessment for training feedback.

**Abstract**

***Background*** Our previous work classified a taxonomy of needle driving gestures during a vesicourethral anastomosis of robotic radical prostatectomy in association with tissue tears and patient outcomes. Herein, we train deep-learning based computer vision (CV) to automate the identification and classification of suturing gestures for needle driving attempts.

***Methods*** Two independent raters manually annotated live suturing video clips to label timepoints and gestures. Identification (2,395 videos) and classification (511 videos) datasets were compiled to train CV models to produce two- and five-class label predictions, respectively. Networks were trained on inputs of raw RGB pixels as well as optical flow for each frame. We explore the effect of different recurrent models (LSTM vs. convLSTM). All models were trained on 80/20 train/test splits.

***Results*** We observe that all models are able to reliably predict either the presence of a gesture (identification, AUC: 0.88) as well as the type of gesture (classification, AUC: 0.87) at significantly above chance levels. For both gesture identification and classification datasets, we observed no effect of recurrent classification model choice on performance.

***Conclusions*** Our results demonstrate CV's ability to recognize features that not only can identify the action of suturing but also distinguish between different classifications of suturing gestures. This demonstrates the potential to utilize deep learning CV towards future automation of surgical skill assessment.

**Introduction**

Growing evidence supports that superior surgical performance is associated with superior clinical outcomes.[1,2] Yet how we presently assess surgery --manual evaluation by peers -- is fraught with subjectivity and is not scalable.[3,4]

Tremendous work has been done already to better assess surgeon performance during robot-assisted surgeries. For example, with suturing, the robotic anastomosis competency evaluation (RACE) has been developed to streamline technical skills assessment with objective criteria for each suturing skill domain[5]. Yet even with such a rubric, manual assessment and feedback of every suture performed by a training surgeon is not feasible. Our group previously deconstructed suturing into a clinically meaningful manner to consist of 3 phases (needle position, needle driving, and suture cinching; Fig 1), and further developed a classification system for suturing gestures to standardize the training and assessment of robot-assisted suturing (Fig 2)[6]. We have demonstrated that surgeon selection of gestures at specific anatomic positions during the vesico-urethral anastomosis (VUA) during the robot-assisted radical prostatectomy (RARP) is linked to surgeon efficiency and clinical outcomes (i.e., tissue tear)[6]. We have also demonstrated that when surgeons are instructed on what specific gesture to utilize during the VUA, they are able to shorten the learning curve for this step of the RARP[7].

Computational approaches have already been tapped towards the goal of recognizing and evaluating surgical gestures. Classical computer vision techniques[8], as well as recurrent models using kinematics[9] have been employed previously with modest success. In recent years, neural networks for extracting information from video data have made tremendous strides.[10,11] Indeed, some groups have started to apply such deep learning approaches to commonly available datasets such as the JIGSAWS suture classification dataset.[12] While these prior works have been largely limited to the well-controlled laboratory environment, live application of computer vision-based *identification* and *classification* of suturing gestures will ultimately determine the real-world utility of such technology.

Herein, we utilize deep learning-based computer vision to 1) *identify* suture needle driving activity during live robot-assisted surgery; 2) *classify* suturing needle driving gestures based on a clinically validated categorization we previously described.

**Methods**

In this study, we set out to characterize commonly used architectures employed in action recognition towards the goal of recognizing and classifying surgical stitches. To undertake this study, we started by generating two complementary datasets for training models from videos of a live VUA during a RARP to identify *when* a suturing gesture is happening (gesture *identification*) and *what* gesture is happening (gesture *classification*). Using annotated video data from a previous study,[6] we generated a dataset of short clips corresponding to moments of "needle driving" (Fig 1b) (positive samples) and short clips corresponding to non-needle driving surgical activity (negative samples). This dataset which we call the "identification dataset" contained 2,395 total video clips (1209 positive; 1186 negative) with an average duration of 12.2 seconds. For gesture classification, we generated a dataset of 511 total clips to distinguish five selected gestures from our established taxonomy (Fig 2). These five were selected based on the adequate sample size per class (Gesture 1 - 150 samples, 2-101, 3-96, 4-117, 5-47). The clips had an average duration of 6.6 seconds and each one was manually labeled by two independent trained annotators. We refer to this dataset as the "classification dataset".

The computational task of identifying actions from video inputs is commonly known in computer vision as *action recognition.* Although a challenging problem, neural networks have recently shown promise in their ability to reason from such spatiotemporal data. The most common example of such networks is so-called "two-stream networks" in which networks take two streams of inputs as features: the *raw RGB pixels* of the video as well as an *optical flow representation* in which momentary direction and magnitude of motion are defined at each pixel (Fig 3). These inputs are usually passed through a standard feature extractor (usually a deep network) and the representations produced by these networks are further passed into a temporally recurrent classification layer, usually some flavor of a long short term memory unit (LSTM[13]). In practice, one can add complexity or inductive biases to the recurrent classification for example by making this layer convolutional (convLSTM[14]), which may aid in performance and training time. In this work, we explore specific hyperparameter choices in this framework for the recurrent classification model (Figure 2). For the comparisons presented here we chose a fixed 7-layer network (AlexNet[15]), which was initialized from weights trained on a large corpus of natural images (ImageNet). We vary the recurrent classification layer (LSTM, convLSTM) in our experiments.

Using the two curated datasets as our starting point, we set out to evaluate commonly used deep learning architectures used in action recognition for the task of identifying *when* (identification) and *what* (classification) suturing gestures happened. Taken together, we hope this work serves as a preliminary demonstration of a potential approach towards merging the latest research in deep learning with the identification, classification, and potential *evaluation* of surgical skills to improve patient outcomes.

### Results

We started by training a model to *identify* short clips as either containing "needle driving" (positive label) or did not contain such an action (negative label) using the annotated identification dataset. We trained all models on three 80/20 train/test splits, using hyperparameters shown in Table 1 and report AUC and accuracy in Figure 4. We observed significantly above chance values for both accuracy (79%) and AUC (0.88) in the identification task, however we found no effect of recurrent classification model on the model performance.

We further moved on to train a model for identifying when a gesture happened using the *classification* dataset to output gesture type probabilities over the 5 selected gestures in our dataset (Figure 2). We varied the same hyperparameters as before (classification layer) and found that similar to the *identification* task that there was no effect of the specific type of classification model. We do however note that convolutional versions of the LSTM (convLSTM) reached convergence in fewer epochs than LSTM counterparts (data not shown). In this classification task, we achieved an average 1st guess (top1) accuracy of 62% for the models trained. Additionally, we also managed to maintain a high AUC (0.87), indicating that the model does not take a biased approach to the classification task to achieve good results. This is further evident in the confusion matrix in Figure 5, where a strong diagonal is present, indicative of reasonable performance in all classes.

## Discussion

In summary, we present a novel annotated dataset for the study of suture gestures in the context of a robot-assisted surgical procedure. We produced annotation for two types of tasks, one with clips annotated with when "needle driving" is present (gesture *identification* dataset) and another dataset labeled with gesture clips and their corresponding type according to the presented taxonomy (gesture *classification* dataset). We further show that applying standard deep network approaches, commonly used in action recognition, can be used to train models that achieve promising performance on both tasks.

The results presented here, in many ways, present a conservative estimate of the sort of performance that can be achieved from these models. We are training in a relatively data-limited regime in both tasks so these models will further improve as labeled data becomes available. In addition, we did not yet employ any inference "tricks" such as ensembling or majority votes commonly used in action recognition models.[10,16]

Our present study is foundational to future work on automating technical skills *evaluation*. Having completed the first steps to identify and classify suturing gestures, we will transition to evaluating how *well* a suture is executed. Part of how well suturing is performed is simply gesture selection at specific anatomic positions[6], in which the present study can help streamline. But the suturing performance also depends on the actual technical skill of the surgeon in carrying out the maneuver, and the models we develop in this study hold promise for such automatic evaluation as well.

On a higher level, our present work is foundational not only for evaluation of suturing, but it also builds the starting point for eventual *autonomous suturing*. Such future platforms must first be capable of recognizing and assessing ideal suturing skills before becoming capable of performing it autonomously.

**COI/Disclosure:** Andrew J. Hung has financial disclyosures with Quantgene, Inc. (consultant), Mimic Technologies, Inc. (consultant), and Johnson & Johnson (consultant).

**Funding/Support:** This study is supported in part by the National Institute Of Biomedical Imaging And Bioengineering of the National Institutes of Health under Award Number K23EB026493.

**Figures**

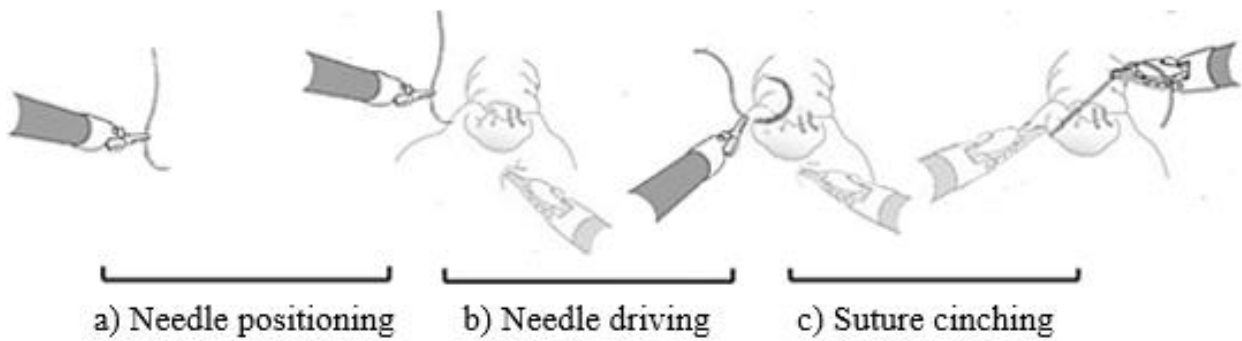

a) Needle positioning    b) Needle driving    c) Suture cinching

**Figure 1. Phases of suturing**

Suturing can generally be broken into 3 repeating phases, including a) "needle positioning" with

needle driver instruments, b) "needle driving" through tissue, c) "suture cinching"

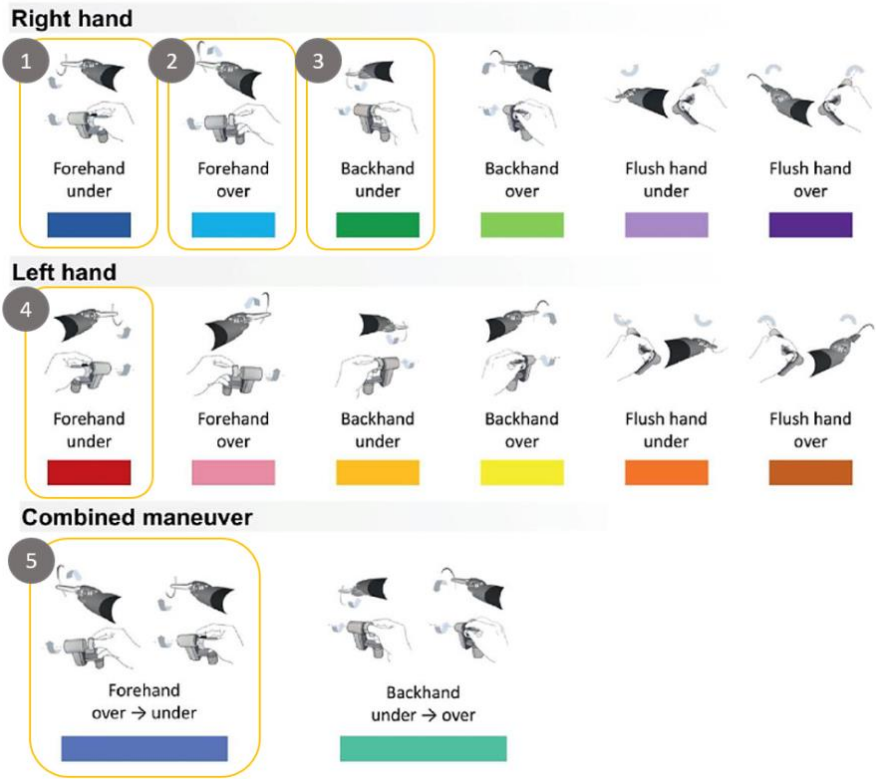

**Figure 2. Taxonomy of suturing gesture types**

The full classification system is presented here, which is derived from our prior work.[6] Boxed gestures refer to those evaluated for our *classification* task in the present study.

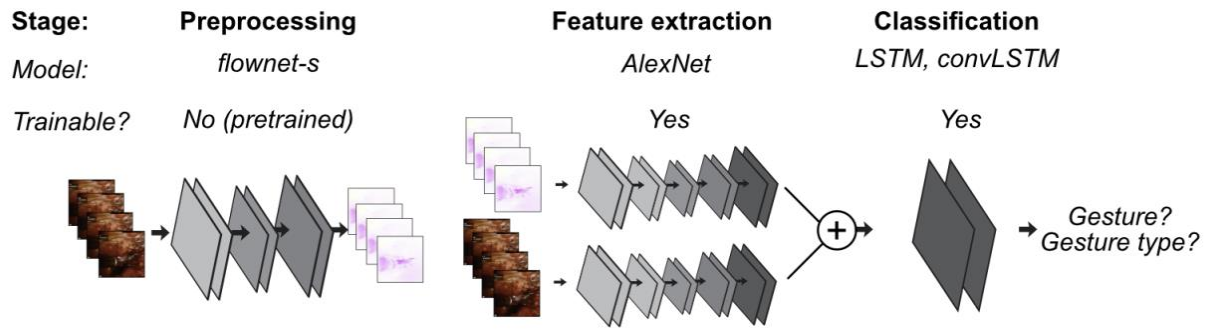

**Figure 3. Data preparation and modeling pipeline**

Schematic of the overall approach to developing a model for gesture presence and gesture type. **Preprocessing:** Prior to applying inputs to any trainable model, we pass the raw RGB video frames through a pre-trained deep network designed to produce optical flow estimates of the video. We code this optical flow into RGB using the hue as direction and saturation as magnitude. We pass this optical flow representation of the video alongside the RGB frames into the subsequent feature extractor networks. **Feature extraction:** We train two feature extractors (one for RGB, one for optical flow) initialized from Imagenet pretrained deep networks. Outputs of these two networks are concatenated before passing to the classification layer.

**Classification:** We train one of two varieties (LSTM, convolutional LSTM) of temporally recurrent classification layers on top of the features extracted. Depending on the task, these models are trained to either produce a 2-class label prediction (gesture *identification*) or a 5-class label prediction (gesture *classification*).

|  | | Gesture identification | | Gesture classification | |
|---|---|---|---|---|---|
| **AlexNet** | LSTM | 0.88 | 0.81 | 0.86 | 0.63 |
| | convLSTM | 0.87 | 0.78 | 0.87 | 0.62 |
| | | **AUC** | **Acc.** | **AUC** | **Acc.** |

**Figure 4. Results summary of identification and classification models on stitching gestures.**

Average model performance across three 80/20 train-test splits of the dataset broken down by task. Models were trained either to predict whether or not a gesture was happening (*identification*) or trained to identify the type of gesture being performed in a clip (*classification*). We vary the recurrent model (LSTM, convLSTM). For the 5-way classification in the gesture classification task, *AUC* represents the average of one vs. rest across classes and *accuracy* represents top1 accuracy.

Confusion matrix AlexNet convLSTM

**Figure 5. Confusion matrix for AlexNet-convLSTM gesture classification model**

Confusion matrix showing normalized accuracy across the 5 possible gesture classes that were produced by the model.

**Table 1: Hyperparameters used during training**

| Hyperparameter | Value (classification) | Value (cutting) |
|---|---|---|
| N classes | 2 | 5 |
| Learning algorithm | Adam | Adam |
| Learning rate | 1e-5 | 1e-5 |
| Epochs | 25 | 7 |
| Batch size | 1 | 1 |
| Base network | {AlexNet} | {AlexNet} |
| Classification network | {LSTM, convLSTM} | {LSTM, convLSTM} |

For each network, 3 different train/test split networks were trained. For video clips longer than 4 sec, a random 4-sec clip was grabbed on each iteration. Classes were balanced via upsampling during training. For data augmentation each frame was resized (240x240) and cropped (224x224) during training. A stride of 4 frames was used (e.g. only sample 1 out of every 4 frames) for an effective frame rate of 7.5Hz. Images were standard scaled before passing through the network. LSTM was a 2-layer LSTM with 64 and 128 units in each layer, respectively. Convolutional LSTM had stride 3 and the number of channels equal to the number of channels in the final convolutional layer of the respective base network (256 for AlexNet).

**Table 2: Number of trainable parameters for each model type**

| | Number of trainable parameters | | |
|---|---|---|---|
| | **Feature extractor** | **Recurrent model** | **Total** |
| AlexNet-LSTM | 3.7 million | 49k | 7.5 million |
| AlexNet-convLSTM | 3.7 million | 7 million | 14.4 million |

Total number of trainable parameters for each of the 2 configurations trained. Note that more parameters does not necessarily improve performance.